\documentclass[10pt]{article}

\usepackage{amsmath,amssymb}

\usepackage{changepage}

\usepackage{textcomp,marvosym}

\usepackage{cite}

\usepackage{nameref,hyperref}

\usepackage{microtype}
\DisableLigatures[f]{encoding = *, family = * }

\usepackage[table]{xcolor}

\usepackage{array}

\newcolumntype{+}{!{\vrule width 2pt}}

\newlength\savedwidth

\raggedright
\usepackage{caption}

\makeatletter
\renewcommand{\@biblabel}[1]{\quad#1.}
\makeatother

\usepackage{lastpage,fancyhdr,graphicx}
\usepackage{bm}
\usepackage{algorithm}
\usepackage{algorithmic}
\usepackage{epstopdf}
\fancyheadoffset[L]{2.25in}
\fancyfootoffset[L]{2.25in}
\begin{document}
\vspace*{0.2in}

\begin{flushleft}
  {\Large
    \textbf\newline{Energy-based survival modelling using harmoniums} 
  }
  \newline
  \\
  H. C. Donker\textsuperscript{1*},
  H. J. M. Groen\textsuperscript{1},
  \\
  \bigskip
  \textbf{1} Department of Pulmonary Diseases, University Medical Center Groningen, Hanzeplein 1, p.o. box 30.001, 9700 RB Groningen, the Netherlands
  \\
  \bigskip

  %
  %

  * h.c.donker@umcg.nl

\end{flushleft}
\section*{Abstract}
Survival analysis concerns the study of timeline data where the event of
interest may remain unobserved (i.e., censored). Studies commonly record
more than one type of event, but conventional survival techniques  focus
on a single event type. We set out to integrate both multiple
independently censored time-to-event variables as well as missing
observations.
An energy-based approach is taken with a bi-partite structure between
latent and visible states, known as harmoniums (or restricted
Boltzmann machines).
The present harmonium is shown, both theoretically and experimentally, to
capture non-linearly separable patterns between distinct time recordings. We
illustrate on real world data that, for a single time-to-event variable,
our model is on par with established methods. In addition, we demonstrate
that discriminative predictions improve by leveraging an extra
time-to-event variable.
In conclusion, multiple time-to-event variables can be successfully captured within
the harmonium paradigm.



\section*{Introduction}

Survival analysis considers the timing of dichotomous events, and can be used to
analyse, for example, time to death, breakdown of a machine, or worsening of the disease. What distinguishes time-to-event
measurements from other random variables is that they are partially observed.
That is to say, not all events have occurred during the interval in which they
were monitored. For example, subjects may be lost during
follow-up, or the experiment may be too short to observe all the events. These
incomplete measurements, where the event time remains unknown, are said to be
censored. Nevertheless, the time interval in which subjects were observed prior
to dropout still provides information. In fact, failure to take into
account censoring leads to serious underestimation of the survival, as has been
repeatedly emphasised~\cite{HARR96, KLEI96, SCHW00, CLAR03a}.

A wide range of statistical tools have been developed to deal with censored
data. These methods rely on modelling the survival distribution $S(\tau)=\int_{\tau}^\infty
  \mathrm{d}t p(t)$ where $p(t)$ is the probability density for observing
the event at time $t$~\cite{KLEI96}. Perhaps the most widely adopted model is
Cox regression~\cite{COX72}. The Cox model makes a proportional hazards (PH) assumption to factorise the hazard function $h(t)\equiv
  p(t)/S(t)$ into a baseline hazard $h_0(t)$ and a (log-) linear function $\exp(\bm{\beta}^T \bm{\bm{x}})$ with weights $\bm{\beta}$ as $h(t)=h_0(t)\exp[\bm{\beta}^T \bm{\bm{x}}]$.

More recently, there are efforts to combine machine learning techniques with
survival analysis. For instance, staying within the PH setting, one can use boosting~\cite{BIND08} to learn the
parameters, or extend the linear function using a neural network architecture,
as done in Refs.~\cite{FARA95, KATZ18, CHIN18}. Neural network structures that
go beyond the PH assumption usually rely on the binning of the time-to-event
variables~\cite{BIGA98, GIUN18, GENS19}. Apart from neural networks, other
models such as random forests~\cite{ISHW08} and support vector
machines~\cite{SHIV07,BELL07, BELL11, POLS15} have been developed as well.

While these models focus on one survival variable, joint analysis of
multiple unordered and distinct time recordings has received comparatively
little attention. A more traditional statistical approach, such as the Wei-Lin-Weissfeld model~\cite{WEI89} (see also
Ref.~\cite{WEI97} for a related review), solves a Cox model for
each individual time-to-event variable and subsequently performs joint inference
on the parameters to determine their significance. MEPSUM~\cite{DEAN14} is a
more recent mixture model where each mixing component fits a discrete time
hazard function.

In this work, a different approach is taken by training
an energy based model on the multiple time-to-event variable likelihood
function~\cite{SCHN05}. Specifically, we consider an unsupervised neural network called harmonium~\cite{SMOL86} (or restricted Boltzmann
machine~\cite{HINT02}, as it is also called), and adapt it to survival analysis.

\section*{Background}\label{subsec:harmonium}
Let us first briefly introduce harmoniums and review some of its quintessential properties. The textbook
harmonium~\cite{GOOD16} consists of binary input states $\bm{x}$
($x_i\in\{0,1\}$ where $i=1 \dots n_v$), binary activations $\bm{h}$
($h_j\in\{0,1\}$ where $j=1 \dots n_h$), and an energy function $E$ that
linearly couples $\bm{x}$ and $\bm{h}$ through a receptive field $\bm{W}$ as
\begin{equation}
  E(\bm{x}, \bm{h})=\bm{x}^T\bm{W}\bm{h}.
\end{equation}
The energy function encodes a
preference for assignments of $\bm{x}$ and $\bm{h}$ that lead to a low $E$.
The probability distribution is parametrised by the energy $E$ as
\begin{equation}
  p(\bm{x}, \bm{h}) = \frac{1}{Z} e^{-E(\bm{x}, \bm{h})} ,
\end{equation}
and a normalisation constant $Z$, called the partition function. Here, the partition function only depends on the free parameters $\bm{W}$. While the latent states $\bm{h}$ are not
observed, they enrich $p(\bm{x}) = \sum_{\bm{h}}p(\bm{x}, \bm{h})$'s
capacity to capture higher-order (i.e., beyond pair-wise) statistics in
the data~\cite{HINT86}. However, the partition function $Z$ is
intractable~\cite{LONG10} and so is $p(\bm{x})$.
Sampling from $p(\bm{x}|\bm{h})$ and $p(\bm{h}|\bm{x})$
is nevertheless easy thanks to the bipartite structure of $E(\bm{x}, \bm{h})$.
The interpretation of
$\bm{W}$ as a receptive field derives from the activation function of $h_j$
given the visible states $\bm{x}$, i.e.,
$p(h_j=1|\bm{x})=\sigma(-\sum_{i=1}^{n_v}x_i W_{ij})$ with sigmoid activation
function $\sigma(x)=1/(1+\exp[-x])$, which is structurally akin to a neural network. A similar relation holds for the binary visible states $\bm{x}|\bm{h}$.

Given a set of $m$ samples
$\{\bm{x}^{(i)}\}_{i=1}^m$, training proceeds by adjusting the free parameters
$\bm{\Theta}$ in $E(\bm{x}, \bm{h})$---which in this case consists of
$\bm{W}$---to maximise the log-likelihood function

\begin{equation}\label{eq:harmonium_likelihood}
  \mathcal{L}(\{\bm{x}^{(i)}\}_{i=1}^m) = \frac{1}{m}\sum_{i=1}^m \ln p(\bm{x}^{(i)}),
\end{equation}

by approximating its gradient using Gibbs samples. The contrastive divergence
algorithm relies on the decomposition of the free parameter $\bm{\Theta}$ gradient of the
likelihood
\begin{equation}
  \nabla_{\bm{\Theta}} \mathcal{L} = -\left(
  \left \langle
  \nabla_{\bm{\Theta}} E
  \right\rangle_{p(\bm{h}|\bm{x})p_{\mathrm{data}}(\bm{x})}
  -
  \left \langle
  \nabla_{\bm{\Theta}} E
  \right\rangle_{p(\bm{x},\bm{h})}
  \right) ,
\end{equation}
into an expectation over the empirical data [first term on the right hand side
    (rhs), $p_\mathrm{data}(\bm{x})=\frac{1}{m} \sum_{i=1}^m
      \delta_{\bm{x},\bm{x}^{(i)}}$] called the \textit{positive phase} and an
expectation over the model itself ($\nabla_{\bm{\Theta}} \ln Z = -\langle
  \nabla_{\bm{\Theta}}E(\bm{x},\bm{h}) \rangle_{p(\bm{x},\bm{h})}$, second term rhs) referred to as the
\textit{negative phase}~\cite{GOOD16}.

While the positive phase can be evaluated in closed form, the negative phase (i.e., the partition function
gradient~\cite{GOOD16} $-\langle \nabla_{\bm{\Theta}}E(\bm{x},\bm{h}) \rangle_{p(\bm{x},\bm{h})}$)
is to be approximated by Gibbs sampling between $p(\bm{x}|\bm{h})$ and
$p(\bm{h}|\bm{x})$. The key empirical observation behind the contrastive
divergence algorithm~\cite{HINT02, HINT12} is that by initialising the chain with training data, a single Gibbs step usually suffices to estimate
the negative phase.

\section*{Theory}\label{sec:theory}
Having briefly reviewed harmoniums, let's now turn to survival
data. We set out to: (i) design an energy function that models survival, categorical, and continuously valued variables, (ii) adapt the
likelihood function to account for censoring and completely
missing data, and (iii) layout a corresponding training algorithm.

\subsection*{Energy function}\label{sec:energy}
\begin{figure}[t]
  \centering
  \includegraphics[width=0.9\columnwidth]{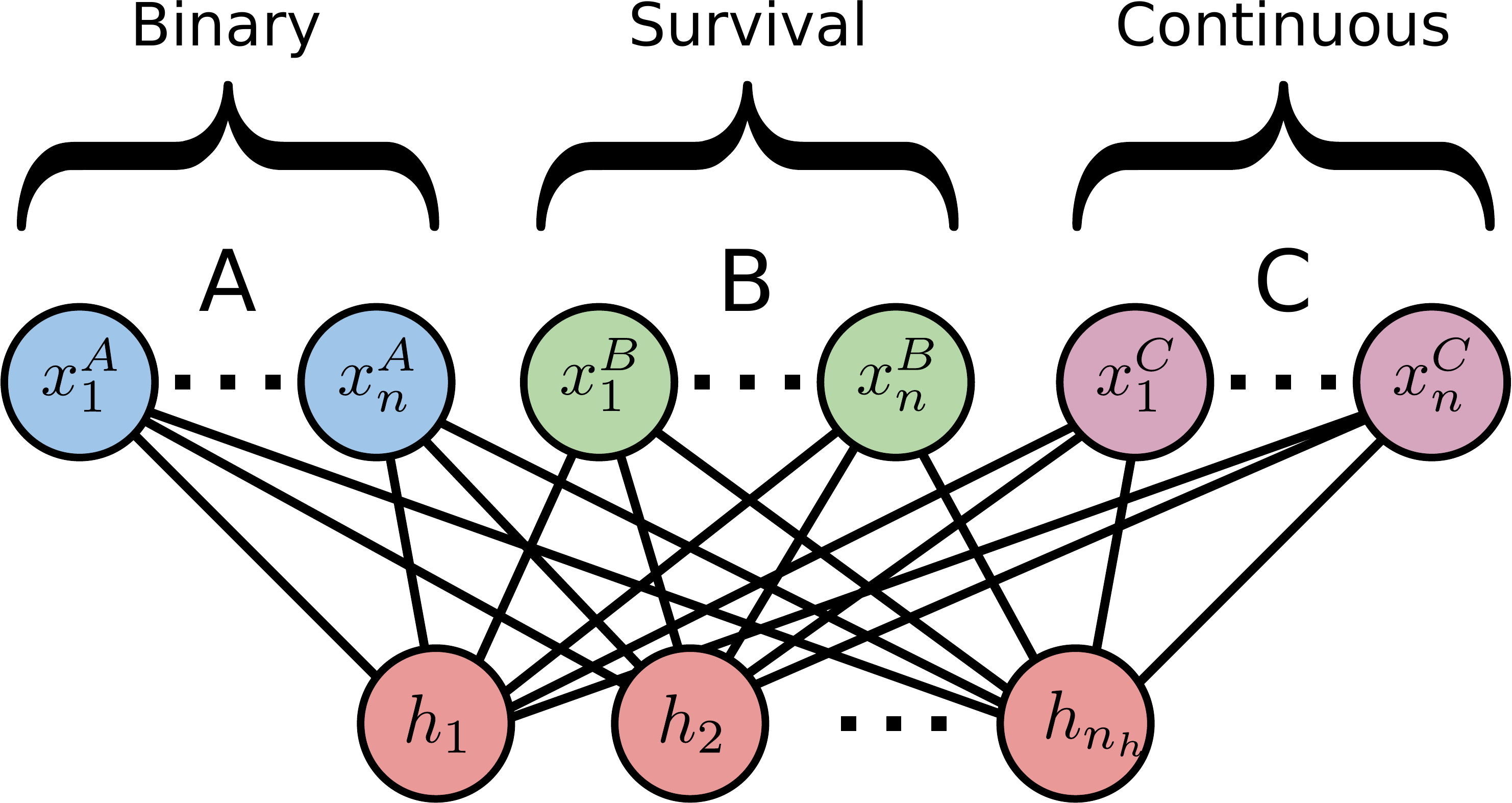}
  \caption{\textbf{Graphical representation of the energy function $E(\bm{x},\bm{h})$.}
    Four types of variables (nodes) can be distinguished: binary states
    $\bm{x}_A$, time-to-event variables $\bm{x}_B$, continuous variables
    $\bm{x}_C$, and (unobserved) binary latent states $\bm{h}$. Edges
    between $\bm{x}$ and $\bm{h}$, indicating the receptive fields, form a
    bipartite graph.}
  \label{fig:harmonium}
\end{figure}

To reiterate, the energy function $E$ codifies preferences for specific
assignments of the variables. Since the event times are continuously valued
instead of binary, we adjust the energy function accordingly. Beside the the survival events, we set out to
capture binary variables (e.g., smoking status) and continuous values (e.g., body
mass index). Let us therefore distinguish between three sets of input variables
(denoted by $A$, $B$, and $C$):
\begin{itemize}
  \item Categorical variables $\bm{x}_A = \{ x_i: i \in A \}$ that are binary encoded $x_{i \in A} \in \{0, 1\}$.
  \item Time-to-event variables $\bm{x}_B = \{x_i: i\in B \}$ that are  scaled to the unit interval $(0,1]$.
  \item Other continuous variables $\bm{x}_C = \{x_i: i\in C\}$ defined on the real line $x_{i \in C} \in \mathcal{R}$.
\end{itemize}
Both the binary variables (in $A$) and the continous variables (in $C$) are assumed
to be independent of time. [See \nameref{S1_Appendix} Sec. 1 for a discussion of the time-to-event variable interval.]
In addition, a \textit{single} set of latent variables $\{h_i\}_{i \in H}$ is used to coherently model the three sets of input
variables. The latent code will be restricted to binary values $h_i \in \{0,1\}$
in view of its regularising effect~\cite{HINT12}.

Assigning an energy term to each variable type gives rise to the overall energy
function
\begin{equation}\label{eq:E}
  E(\bm{x}, \bm{h}) = E_A + E_B + E_C + E_H.
\end{equation}

For the categoric $\bm{x}_A$ and continuous $\bm{x}_C$ variables we rely on established energy functions:
(i) $E_A$ is modelled as a binary-binary harmonium~\cite{HINT12} including a bias term,
(ii) $E_C$ represents a Gaussian-binary harmonium ~\cite{THEI11, HINT12,
  MELC17}.
For the time-to-event variables a new function $E_B$ is proposed (see Sec. 1, \nameref{S1_Appendix}) for which $p(x_i|\bm{h})$ is
a (truncated) gamma distribution. The term $E_H$ contains a bias for the latent states $\bm{h}$.
Intuitively, the bi-partite structure of these energy terms control the conditional distributions $p(\bm{x}|\bm{h})$.
In turn, the conditional distributions can be seen as the building blocks of the model with weight $p(\bm{h})$.
By training the parameters $\bm{\Theta}$ we adjust the weights $p(\bm{h})$ of the blocks to refine the fit.
More concretely, we take the following energy functions

\begin{eqnarray}\label{eq:E_ABCH}
  E_A & = & \bm{x}_A^T \bm{W}_A \bm{h} + \bm{x}_A^T \bm{a}_A \, , \\
  E_B & = &  \bm{x}_B^T \bm{W}_B \bm{h}  - \ln(\bm{x}_B)^T |\bm{V}| \bm{h} \nonumber \\
  & & +  \bm{x}_B^T \bm{a}_B - \ln (\bm{x}_B)^T|\bm{c}| \, , \\
  E_C & = & (\bm{x}_C \oslash \bm{\sigma})^T \bm{W}_C \bm{h}  + \frac{1}{2} \left\|(\bm{x}_C - \bm{a}_C)\oslash \bm{\sigma} \right\|^2 , \\
  E_H & = & \bm{b}^T \bm{h} \, ,
\end{eqnarray}

where $\oslash$ is the Hadamard division operator and the absolute value $|\cdot|$ is applied element wise.
The weights $\{\bm{W}_A, \bm{W}_B, \bm{W}_C\}$ can be interpreted as the
receptive fields of $\bm{x}$ to activate the latent states $\bm{h}$, while
$\{\bm{a}_A, \bm{a}_B, \bm{a}_C\}$ and $\bm{b}$ are their respective biases. The
$\bm{V}$ and $\bm{c}$ terms in $E_B$ are additional receptive fields and biases
that help modulate the survival distribution. The receptive fields of
$E(\bm{x},\bm{h})$, coupling $\bm{x}$ and $\bm{h}$, are illustrated in
Fig~\ref{fig:harmonium} by corresponding edges. The form of
$E(\bm{x}, \bm{h})$ fixes the distribution over $\bm{x}$ given $\bm{h}$ leading to
\begin{equation}\label{eq:p_x_cond_h}
  p(x_i|\bm{h}) = \left\{
  \begin{matrix}
    \sigma[(1-2x_i)z_i]                & i \in A, \\
    p_\Gamma(x_i|\alpha_i, \beta_i)    & i \in B, \\
    \mathcal{N}(x_i|\mu_i, \sigma_i^2) & i \in C,
  \end{matrix}
  \right.
\end{equation}
where $\sigma[(1-2x_i)z_i]$ is the sigmoid function with latent state activation $\bm{z} = \bm{a}_A + \bm{W}_A \bm{h}$.
The right truncated Gamma distribution $p_\Gamma(x_i|\alpha_i, \beta_i)$ [Eq.~(2), \nameref{S1_Appendix}] has
shape $\bm{\alpha} = |\bm{V}|\bm{h} + |\bm{c}| + \bm{1}$ (elementwise absolute value)
and rate $\bm{\beta}
  = \bm{W}_B \bm{h} + \bm{a}_B $. Finally, $\mathcal{N}(x_i|\mu_i,
  \sigma_i^2)$ is a Gaussian with mean $\bm{\mu} = \bm{a}_C - \bm{\sigma}\circ (\bm{W}_C \bm{h})$ (with $\circ$ denoting Hadamard product) and standard deviation $\bm{\sigma}$. In a similar way, the
activations of the latent variables
\begin{equation}\label{eq:p_h_cond_x}
  p(h_j|\bm{x}) = \sigma[(1-2h_j)\phi_j],
\end{equation}
depend on the contributions of all variable types, which are jointly captured by
\begin{eqnarray}\label{eq:phi}
  \bm{\phi} & = & \bm{b} +  {\bm{W}_A}^T\bm{x}_A + {\bm{W}_C}^T(\bm{x}_C \oslash \bm{\sigma}) \nonumber\\
  & & + {\bm{W}_B}^T \bm{x}_B + |\bm{V}^T| \ln \bm{x}_B.
\end{eqnarray}
A key observation that is central to the training of harmoniums is that Gibbs samples from
$p(\bm{x},\bm{h})$ can be obtained by alternating between
Eq.~(\ref{eq:p_x_cond_h}) and Eq.~(\ref{eq:p_h_cond_x}). In this way, an entire
block of states can be updated in parallel, thanks to its conditional
independence.

\subsection*{Cost objective}\label{sec:objective}
Next, we adapt the likelihood function to incorporate partially and completely missing values. We will
assume independent and uninformative censoring and that missing values are missing at random. To
simplify the exposition we focus on right censored data (or
censored, for short). That is, observations for which there is a lower
bound on the failure time (e.g., a participant that was lost to follow-up after
time  $t$).
The standard likelihood approach for modelling a single
time-to-event variable is as follows.
When a sample is censored at time $t$, we replace $p(t)$ by its survival function $S(t)$.
Writing $S(t) = \int_0^\infty \mathrm{d}\tau
  \Theta(\tau-t) p(\tau)$ with the Heaviside step function $\Theta(x)= \left \{
  \begin{matrix}1 & x \geq 0 \\ 0 & x < 0\end{matrix} \right.,$
shows that this corresponds to marginalising out the unobserved region.
Both cases, censored and
observed events, can be succinctly codified as an integration over the domain of
event times $\int_{0}^{\infty} \mathrm{d}\tau  p(\tau) \chi(t, \tau, e)$
constrained by $ \chi(t, \tau, e) = \delta(\tau - t)^e \Theta(\tau - t)^{1-e}$
with $\delta(x)$ the Dirac delta function and  $e=1$ ($e=0$) indicating
observation (censoring) at time $t$. Note the analogy with completely missing
data where the entire domain (instead of subset of the domain) is marginalised,
e.g., $p(t_1)=\int_{-\infty}^{\infty}\mathrm{d}t_2 p(t_1,t_2)$ when $t_2\in
  \mathcal{R}$ is missing. We can therefore apply a similar codification scheme to
missing values to obtain $\chi(t, \tau, e)=\delta(\tau-t)^e$. A consistent
generalisation from one to multiple censored variables is straightforward: integrate out
the entire unobserved region~\cite{SCHN05}.
More precisely, let
$e_a=1$ indicate the occurrence and $e_a=0$ the absence of observation $x_a$.
That is, $e_a=0$ indicates that $x_a$ is censored when $a$ refers to a
time-to-event variable ($a\in B$), or completely missing otherwise ($a \in A
  \cup C$). In addition, let $\xi_a$ be the corresponding observed value when
$e_a=1$, its lower bound (i.e., censoring time) for the survival variables
(i.e., $a\in B$) or a placeholder $\xi_a=?$ otherwise ($a \in A \cup C$) when
$e_a=0$. As a shorthand, denote $o_a=(\xi_a, e_a)$ and a superscript $o_a^{(i)}$
to refer to a specific sample $i$. First, group the marginalisation constraints
that are imposed by the observations

\begin{equation}\label{eq:constraint}
  \chi(\bm{x}, \bm{o}) = \prod_{i\in A\cup B \cup C} \delta(x_i - \xi_i)^{e_i} \prod_{j \in B} \Theta(x_j -  \xi_j)^{1 - e_j} ,
\end{equation}
with $\Theta(x)$ the Heaviside step function and $\delta(x)$ the Dirac delta function (Kronecker delta function) for the continuous variables in $B \cup C$ (binary variables in $A$). Equation (\ref{eq:constraint}) is a symbolic way to represent that we should either pick the observed values (the delta function) or marginalise the unobserved region (which, for the survival variables, is the interval starting from the censor time, or the entire domain otherwise). In this way, the likelihood can be expressed as
\begin{equation}\label{eq:likelihood_integral}
  L(\{\bm{o}^{(i)}\}_{i=1}^m) = \prod_{i=1}^m \int\mathrm{d}\bm{x} p(\bm{x}) \chi(\bm{x}, \bm{o}^{(i)}),
\end{equation}
using the shorthand
$\int \mathrm{d}\bm{x} \equiv \int_{-\infty}^{\infty} \mathrm{d}\bm{x}_C \int_0^1 \mathrm{d}\bm{x}_B \sum_{\bm{x}_A \in  \{0, 1\}\otimes^{|A|}}$.
In summary, to train the model that takes into account
censored and missing data should strive to optimise the likelihood function
Eq.~(\ref{eq:likelihood_integral}) or, equivalently, the log-likelihood function
$\mathcal{L}(\{\bm{o}^{(i)}\}_{i=1}^m)=\ln(L)/m$.

Having spelled out the likelihood function in fair generality, next we apply it
to the energy parameterisation $p(\bm{x}) \propto \sum_{\bm{h}} \exp[-E(\bm{x}, \bm{h})]$.
To keep the bi-partite structure intact
we turn to a trick from Ref.~\cite{TRAN13}
to reformulate the model in terms of $p(\bm{o}, \bm{x}, \bm{h})$~\cite{TRAN13},
where
\begin{equation}\label{eq:tran}
  p(\bm{o}, \bm{x}, \bm{h}) \propto e^{-E(\bm{x}, \bm{h})}\chi(\bm{x}, \bm{o}).
\end{equation}
With the help of Eq.~(\ref{eq:tran}) the gradient of the log-likelihood (details are in Sec. 2, \nameref{S1_Appendix}) can be
expressed as
\begin{equation}\label{eq:gradient_likelihood}
  \nabla_{\bm{\Theta}} \mathcal{L} = -\left( \left \langle \nabla_{\bm{\Theta}}E \right\rangle_{p(\bm{x}, \bm{h}|\bm{o})p_{\mathrm{data}}(\bm{o})} - \langle \nabla_{\bm{\Theta}}E \rangle_{p(\bm{x}, \bm{h})} \right),
\end{equation}
where $p_{\mathrm{data}}(\bm{o}) = \frac{1}{m}\sum_{i=1}^m\delta_{\bm{o}, \bm{o}^{(i)}}$. Heuristically speaking, Eq.~(\ref{eq:gradient_likelihood}) indicates that the
gradient contrasts the the empirical statistics of
$\nabla_{\bm{\Theta}}E(\bm{x}, \bm{h})$ incorporating the constraints [through $p(\bm{x},
      \bm{h}|\bm{o})$, first term, rhs] with the models own perception [generated by $p(\bm{x},\bm{h})$] of
$\nabla_{\bm{\Theta}}E(\bm{x}, \bm{h})$ (second term, rhs).

\subsection*{Training}\label{sec:training}
\begin{algorithm}[t]
  \begin{algorithmic}[1]
    \WHILE{not converged}

    \STATE{Load minibatch $\{\bm{o}^{(1)}, \dots,\bm{o}^{(m)}\}$ of $m$ samples.}

    \FOR{$i=1$ to $m$}
    \STATE $\bm{x}^{(i)} \gets \bm{\xi}^{(i)}$
    \STATE $\tilde{\bm{x}}^{(i)} \gets \bm{\xi}^{(i)}$
    \FOR{$l=1$ to $k$}
    \STATE $\bm{h}^{(i)} \sim p(\bm{h}^{(i)}|\bm{x}^{(i)})$
    \COMMENT{Sample positive phase.}
    \STATE $\bm{x}^{(i)} \sim p(\bm{x}^{(i)}|\bm{h}^{(i)},\bm{o}^{(i)}) $
    \STATE $\tilde{\bm{h}}^{(i)} \sim p(\tilde{\bm{h}}^{(i)}|\tilde{\bm{x}}^{(i)})$
    \COMMENT{Sample negative phase.}
    \STATE $\tilde{\bm{x}}^{(i)} \sim p(\tilde{\bm{x}}^{(i)}|\tilde{\bm{h}}^{(i)}) $
    \ENDFOR
    \STATE $\bm{\mu}^{(i)} \gets p(\bm{h}^{(i)}=\bm{1}|\bm{x}^{(i)})$
    \STATE $\tilde{\bm{\mu}}^{(i)} \gets p(\tilde{\bm{h}}^{(i)}=\bm{1}|\tilde{\bm{x}}^{(i)})$
    \ENDFOR
    \STATE \COMMENT{Gradient ascent update with learning rate $r_{\mathrm{learn}}$.}
    \STATE $\Delta \bm{\Theta} \gets -\sum_{i=1}^m \frac{\nabla_{\bm{\Theta}}E(\bm{x}^{(i)},\bm{\mu}^{(i)}) - \nabla_{\bm{\Theta}}E(\tilde{\bm{x}}^{(i)},\tilde{\bm{\mu}}^{(i)}) }{m}$.
    \STATE $\bm{\Theta} \gets \bm{\Theta} + r_{\mathrm{learn}} \Delta \bm{\Theta}$.
    \ENDWHILE
  \end{algorithmic}
  \caption{The $k$-step contrastive divergence algorithm for censored and missing values.}
  \label{alg:training}
\end{algorithm}

Next, we discuss how to maximise the likelihood with
gradient ascent by approximating the gradient
  [Eq.~(\ref{eq:gradient_likelihood})]. In the standard contrastive divergence~\cite{HINT02} approach,
the negative phase is approximated using Gibbs samples while the positive phase
can be evaluated exactly. Incorporating missing and censored values has modified
the positive phase [first term, rhs Eq.~(\ref{eq:gradient_likelihood})] in a way
that evades a closed form solution. Instead, Eq.~(\ref{eq:gradient_likelihood})
will be estimated by Gibbs sampling both the positive and the negative phase.

Analogous to the sampling of $p(\bm{x},\bm{h})$ for the negative phase, we alternate
between $p(\bm{x}|\bm{o}, \bm{h})$ and $p(\bm{h}|\bm{o}, \bm{x})$ to generate samples
of $p(\bm{x},\bm{h}|\bm{o})$ for the positive phase where

\begingroup
\setlength\arraycolsep{1pt}

\begin{eqnarray}\label{eq:p_x_cond_o_h}
  p(x_i|\bm{o}, \bm{h}) = \left\{
  \begin{matrix}
    \delta(x_i - \xi_i) & \forall i      & e_i =1,  \\
    p(x_i| \bm{h})      & i \in A \cup C & e_i = 0, \\
    p^{\Gamma}_{[\xi_i, 1]}(x_{i}|\alpha_i, \beta_i )
                        & i \in B        & e_i = 0, \\
  \end{matrix}
  \right.
\end{eqnarray}
\endgroup
with $p^{\Gamma}_{[\xi_i, 1]}\left[x_{i}|\alpha_i(\bm{h}), \beta_i(\bm{h})\right]$ the gamma distribution
normalised to the $[\xi_i, 1]$ interval [Eq.~(24), \nameref{S1_Appendix}] and
\begin{equation}
  p(\bm{h}|\bm{o}, \bm{x}) = p(\bm{h}|\bm{x}).
\end{equation}
Physically, Eq.~(\ref{eq:p_x_cond_o_h}) indicates that samples $x_i$ should
adhere to the bounds imposed by the observation (which is the lower-bound censor
time for the censored variables).

In summary, the training procedure is as follows: For the positive phase [first
    term r.h.s. Eq.~(\ref{eq:gradient_likelihood})] replace the censored and missing
values with ``\textit{fantasy states}'' and calculate the
$\nabla_{\bm{\Theta}}E(\bm{x}, \bm{h})$ statistic. To replace the censored data, clamp the
observed events and sample the censored events from the unobserved interval and
sample the missing values from the entire distribution
  [Eq.~(\ref{eq:p_x_cond_o_h})]. The negative phase [second term r.h.s.
    Eq.~(\ref{eq:gradient_likelihood})] is calculated similarly, but now all the
states are updated (none are clamped) from the entire interval
  [Eq.~(\ref{eq:p_x_cond_h})]. To generate fantasy states, pick the the mini batch
as the initial state of the Gibbs chain, initialise the placeholder ``$?$'' by
the median value over the training set, and carry out $k$ Gibbs chain steps.
Finally, update the weights using the phase difference and repeat the entire
process until some predefined stopping criterion. In pseudocode, the algorithm
is outlined in Algorithm~\ref{alg:training}. Reassuringly, the original
contrastive divergence algorithm is recovered as a special case when all the
data is observed (i.e., $e^{(i)}_a=1$ for all $i$ and $a$).

\section*{Example: a three-way problem}\label{sec:example}
\begin{figure*}[t]
  \centering
  \includegraphics[width=0.8\textwidth]{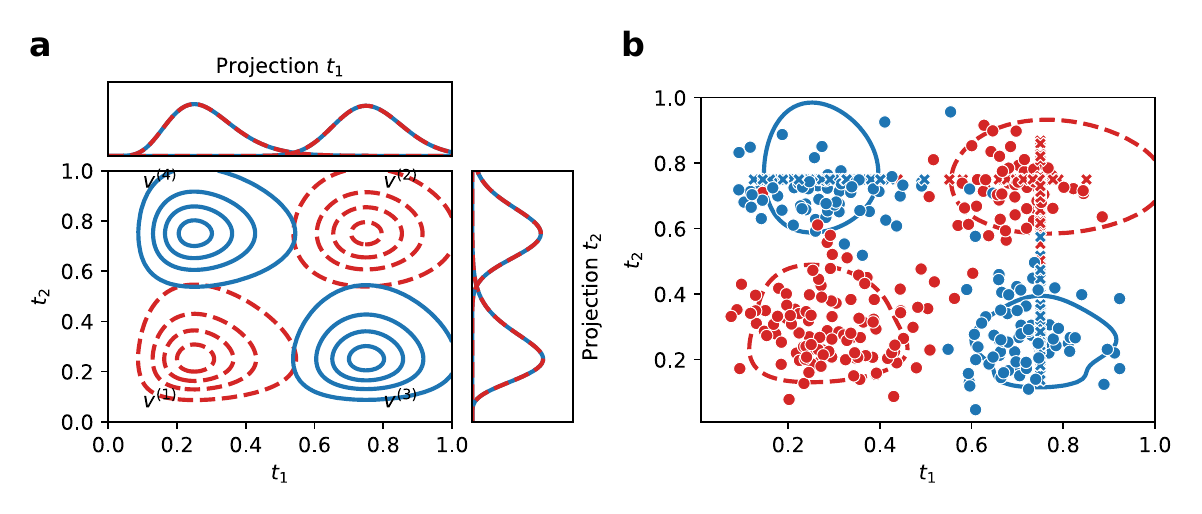}
  \caption{\textbf{A harmonium captures non-linearly seperable time recordings.}
    (\textbf{a}) A synthetic
    time-to-event distribution where the timing of two separate events,
    $t_1$ and $t_2$, are correlated or anti-correlated according to red or blue colour, respectively. The density is
    composed of two red ($v^{(1)}$ and $v^{(2)}$, dashed contours) and two
    blue  ($v^{(3)}$ and $v^{(4)}$, solid contours) blobs.
    Side panels show the marginal probability density by colour,
    illustrating the multivariate nature of the problem. (\textbf{b}) Model fit (contours) of
    observed (dots) and censored recordings (crosses) sampled from (\textbf{a}). The contours,
    indicating constant probability density of the harmonium,
    shows that all four colour-mode combinations are recapitulated.
  }
  \label{fig:xor_example}
\end{figure*}

The limitations of uni-survival variate models (i.e., conventional survival analysis) is
best illustrated with a three-way problem. Consider a distribution generating two event recordings $\bm{x}_B=[t_1, t_2]^T$ and
a colour, red ($x_A=0$, we've dropped the index for convenience) or blue ($x_A=1$).
Let the probability density be confined
to the unit square $[0, 1] \times [0, 1]$ symmetrically tiled with four equally
weighted bell-shaped blobs. Anti-correlated recordings (blue) along the diagonal, and correlated recordings (red) on the off-diagonal quadrants (see Fig~\ref{fig:xor_example}a and Sec.~3, \nameref{S1_Appendix} for details).

Looking at the projections (i.e., marginals) along the axes (side panels
Fig~\ref{fig:xor_example}a), shows how the red and blue modes collapse onto
each other. Viewed from either $t_1$ or $t_2$ alone, one would therefore be
inclined to (falsely) conclude there is no relation between colour and survival.

For a harmonium with $n_h=4$ hidden units we can derive a closed-form approximate solution for this three-way problem.
The solution approximates a mixture of Gaussians in survival space
(parametrised by $\bm{x}_B$) with $x_A$ clamped to mode $j$'s colour
which we call ${(\tilde{x}_A)_j}$ (see Sec.~4 \nameref{S1_Appendix}, for a derivation).
The probability density can be approximated as

\begin{eqnarray}
  p(x_A, \bm{x}_B) & = & \frac{1}{Z} \sum_{h_1,\dots,h_4} \exp[-E(x_A, \bm{x}_B, \bm{h})] \nonumber\\
  & \approx & \frac{1}{4}\sum_{j=1}^4 \delta_{x_A,(\tilde{x}_A)_j} \mathcal{N}\left(\bm{x}_B|\bm{v}^{(j)}, \bm{\Sigma}_j\right),
\end{eqnarray}

where the mean $\bm{v}^{(j)}$ and the (diagonal) covariance matrix $\bm{\Sigma}_j$ are determined through the
rows of the receptive fields $\bm{v}^{(j)} = \frac{|\bm{V}_{j}|}{(\bm{W}_B)_{j}}$ and
$\bm{\Sigma}_j=\frac{|\bm{V}_{j}|}{{(\bm{W}_B)_{j}}^2}$, with all other visible
biases zero (all other weights are described in Sec.~4 \nameref{S1_Appendix}).

Knowing that the problem is solvable in theory, lets illustrate
training with censored time recordings. We generated 1000 samples and censored each event
time $t_i > \frac{3}{4}$ with 75 \% probability. For clarity, half of the points
are shown in Fig~\ref{fig:xor_example}b, coloured by $x_A$, and marked by a cross
where censored. The harmonium was trained for $3 \cdot 10^5$ epochs with a
learning rate $r_{\mathrm{learn}}=0.375$, 10 \% momentum, and 3
persistent~\cite{TIEL08} contrastive divergence sampling steps.

While a model that doesn't account for censoring would underestimate survival,
we find that the harmonium correctly identifies all four modes. We do observe
that the modes are less sharply peaked (more smeared) compared to the original
distribution. This is attributed to the approximate and stochastic nature of the
contrastive divergence algorithm, which sometimes hinders convergence. Overall,
the harmonium satisfactory captures the three-point correlation in the survival
data. For reference, we trained a Cox model~\cite{COX72} on either $t_1$ or
$t_2$ with $x_A$ as a covariate. In both cases we found that its regression
coefficient is zero (null hypothesis) under a p-value threshold of 0.05. That
is, the Cox model finds no relation between $x_A$ and survival.

\section*{Experiments}\label{sec:results}
To illustrate performance on real world datasets, the harmonium is compared to
(i) Cox regression~\cite{COX72} from the lifelines package~\cite{DAVI20} with
both $L_1$ and $L_2$ regularisation, (ii) random survival forest~\cite{ISHW08}
and (iii) the fast support vector machine (SVM)~\cite{POLS15}, where the latter
two are both from the scikit-survival package.

\subsection*{Datasets}
Our benchmark is comprised of four lifelines datasets~\cite{DAVI20}, namely:
\begin{itemize}
  \item The recidivism of convicts released from the Maryland state prisons
        ($m$=432 convicts)~\cite{ROSS80}---denoted as \textit{arrest}---to study the
        effect of financial aid.

  \item The duration of democratic and dictatorial political regimes ($m$=1808
        countries)~\cite{CHEI10}---denoted as \textit{democracy}

  \item The survival of women with breast cancer ($m$=686
        patients)~\cite{SCHU94} (denoted as \textit{gbsg2}) to measure the effect of
        hormonal therapy.

  \item The survival of advanced lung cancer patients ($m$=288
        patients)~\cite{LOPR94}---denoted as \textit{ncctg}---where the prognostic
        value of a patient's questionnaire was examined.
\end{itemize}

In addition, our benchmark comprises two additional lung cancer datasets containing two (instead of one) time-to-event recordings (bundled with the code, but not part of lifelines):
\begin{itemize}
  \item A Dutch study, \textit{nvalt11}, considered the effect of profylactic
        brain radiation versus observation in ($m$=174) patients with advanced
        non-small cell lung cancer~\cite{RUYS18}. The \textit{nvalt11} dataset
        contained time recordings for overall survival (OS) and symptomatic brain
        metastasis-free survival (SBMFS).

  \item Another Dutch study, called \textit{nvalt8} ($m$=200 patients), that
        examined if nadroparin combined with chemotherapy could reduce cancer relapse
        after surgical removal of a non-small cell lung tumour~\cite{GROEN19}. The
        dataset contained failure times for both OS and recurrence free survival
        (RFS).
\end{itemize}

\subsection*{Results}
To reiterate, the primary difference between the  harmonium, and the
implementations of Cox model, random survival forest, and SVM considered here,
is that the harmonium can incorporate missing values and multiple (potentially non-linearly
related) survival variables. Results are compared using two metrics: Harrell's
concordance index~\cite{HARR96} and Brier's calibration loss~\cite{GRAF99} at $t=\tau_{\mathrm{OS}}/2$, where the time horizon $\bm{\tau}$ was set to the largest time recording in the dataset.

Since the harmonium captures both survival recordings of the \textit{nvalt}
datasets and computes the concordance and calibration based on the overall survival distribution, we present results both with and without factoring
in the second survival variable. The former are derived from $S(x_{\mathrm{OS}}=t|\bm{o}_{-\mathrm{OS}}) = p\left(x_{\mathrm{OS}} > t,\bm{o}_{-\mathrm{OS}}\right)/p(\bm{o}_{-\mathrm{OS}})$
where $\bm{o}_{-\mathrm{OS}}$ denotes observation $\bm{o}$ with the element indexed by $\mathrm{OS}$ removed (but still containing the other survival variable).
For the latter, the dependence of the other survival variable (SBMFS and RFS) was marginalised out $S(x_{\mathrm{OS}}=t|\bm{o}_{-\{\mathrm{OS},
  \mathrm{SBMFS}\}}) = p\left(x_{\mathrm{OS}} > t,\bm{o}_{-\{\mathrm{OS},\mathrm{SBMFS}\}}\right)/p(\bm{o}_{-\{\mathrm{OS}, \mathrm{SBMFS}\}})$ and similarly $S(x_{\mathrm{OS}}=t|\bm{o}_{-\{\mathrm{OS},
  \mathrm{RFS}\}})$ for \textit{nvalt11} and \textit{nvalt8} and corresponding $t$, respectively. That is, the model does not have access to the additional time-to-event variable during inference (only during the training phase).

\begin{figure*}[t]
  \centering
  \includegraphics[width=0.8\textwidth]{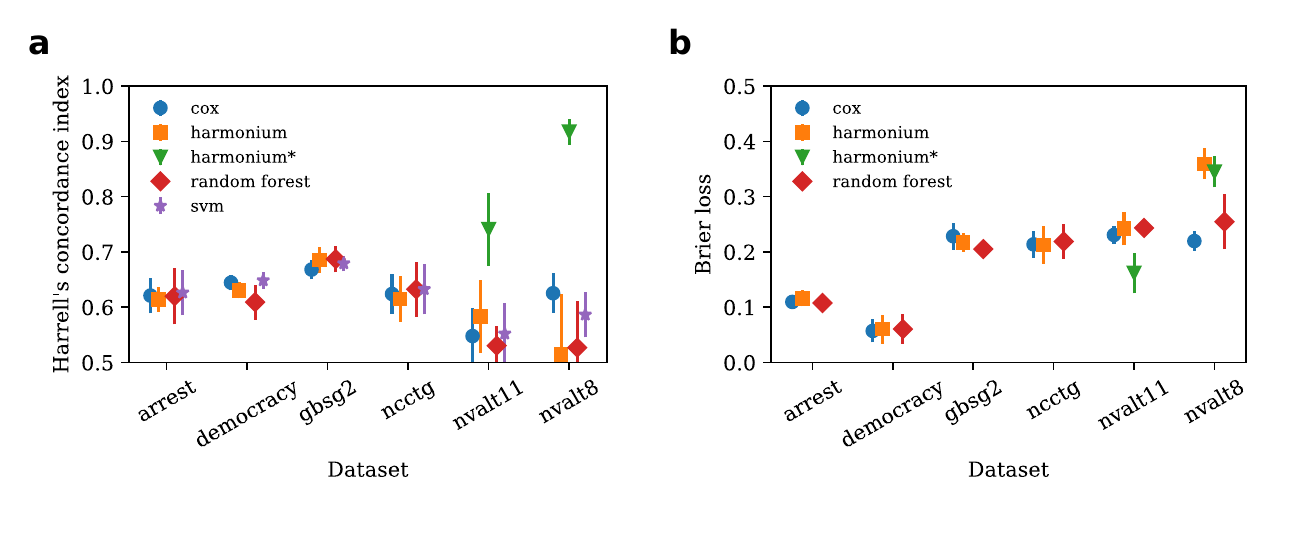}
  \caption{\textbf{A benchmark of survival models (indicated in the legend) across
    various datasets (horizontal axis) shows that the harmonium is on par
    with uni-survival variate models, and discriminative predictions improve
    with an additional survival variable (\textit{nvalt11} and \textit{nvalt8}).}
  Specically, overall survival (OS) and symptomatic brain metastasis-free survival
  (SBMFS), $\bm{x}_B=[x_{\mathrm{OS}}, x_{\mathrm{SBMFS}}]^T$,  were recorded
  for the \textit{nvalt11} dataset; OS and recurrence free survival (RFS),
  $\bm{x}_B=[x_{\mathrm{OS}}, x_{\mathrm{RFS}}]^T$, were recorded for the
  \textit{nvalt8} dataset.
  Metrics for these two datasets were computed for OS but we distinguish
  between two methods of computation for the harmonium. Namely, metrics that
  factor in the second survival variable [indicated by a * in the legend, and
  computed through survival distribution
  $S(x_{\mathrm{OS}}=t|\bm{o}_{-\mathrm{OS}})$] versus the metrics where this
  variable was marginalised out (without a *, via
  $S(x_{\mathrm{OS}}=t|\bm{o}_{-\{\mathrm{OS}, \mathrm{SBMFS}/\mathrm{RFS}\}})$) evaluated at half the overall survival time horizon $t=\tau_{\mathrm{OS}}/2$ . Calibration data was not
  available for the support vector machine (svm). Markers and errorbars
  indicate the mean value and the standard deviation from 5x5 nested
  cross-validation.
  }
  \label{fig:benchmark}
\end{figure*}

The benchmark results are summarised in Fig~\ref{fig:benchmark}. For the \textit{nvalt8} and \textit{nvalt11}
datasets, notice that when we factor in the additional survival information
(indicated by a * in the legend) we observe a substantial improvement in the concordance index (Fig~\ref{fig:benchmark}a). Conversely, when the model did not have access to the extra time-to-event variable during inference (without a *) the performance reduces to that of the other models. 
These results are in line with common
sense: a disease relapse or finding a brain tumour decreases one's expected
life expectancy. Moreover, the performance reduction upon marginalisation further highlights the relation between the two endpoints.
In terms of calibration (Fig~\ref{fig:benchmark}b), the additional survival information leads to a further improvement in the \textit{nvalt11} dataset but not in the \textit{nvalt8} dataset (where it performed slightly worse). For the four other datasets (to wit, \textit{arrest}, \textit{democracy}, \textit{gbsg2}, and \textit{ncctg}), the harmonium performed comparable to other methods in terms of concordance (Fig~\ref{fig:benchmark}a) and calibration (Fig~\ref{fig:benchmark}b). 
Including variables with missing values, as we did for the \textit{nvalt}
datasets, showed no noticeable improvement for the harmonium compared to the
other models (where this could not be taken into account).

\section*{Discussion}\label{sec:discussion_conclusion}
Healthcare data follows an
inherent timeline where new information, such as a lab result or a diagnosis, comes in
continuously. At the same time, most (but not all) statistical and machine learning models require
data in tabular format. This poses a challenge, where one should strike a balance
between a format that accommodates the model and simultaneously does justice to the time
ordering of the data. Our work is a step towards a consolidation of these two
representations of the data, by modelling both missing values and multiple time-to-event
variables in one coherent framework. In contrast to, e.g., competing risks, where one
event excludes another, we (i) require that time recordings are censored independently
but (ii) do not impose \textit{a priori} (e.g., causal) dependence between the survival
variables. Rather, the survival distributions are independent conditional on a latent
variable similar to frailty models of clustered data~\cite{BALA20}. Different from
frailty models, (i) our latent state $\bm{h}$ is a binary vector instead of a continuous
value and (ii) we not assume that the survival distributions are identical given
$\bm{h}$. As a result, not only can we capture anti-correlations, unlike frailty
models~\cite{BALA20}. We can also accomodate three-way correlations, as demonstrated
using the three-way example.

One disadvantage of our model --- like all neural networks --- is the myriad of hyperparameters to tune. Choosing appropriate
parameters for the learning rate, batch size, number of latent states, how many
epochs to train, and regularisation can be challenging. In addition, while some
quantities, e.g., the latent states, can be computed efficiently (i.e., linear
in the number of input variables), others such as the survival distribution
  [Eq.~(23), \nameref{S1_Appendix}] are more computationally demanding. This was why we used the Brier loss instead of the integrated Brier loss.

A second limitation of this work, unrelated to our model, is that Harrell's
concordance index and Brier loss --- both intrinsically uni-survival variate
metrics --- may not be the most appropriate measures to comprehensively interrogate
a model's capacity to capture multiple time recordings. We could only indirectly
probe its performance by conditioning on, and marginalising out, the second survival variable.
Alas, as far as we know, no higher dimensional
generalisations of, e.g., Harrell's concordance index or the Brier loss exist.
In this regard, we
believe that our simple three-way problem can serve as a useful
litmus test for future multi-survival variate models.

\section*{Conclusion}
In conclusion, a new harmonium was proposed for partially and completely missing
data. Multiple distinct time recordings are jointly modelled without imposing
\textit{a priori} relations between events, in contrast to
conventional survival techniques. In addition, time-independent features with missing values can be straightforwardly incorporated
thanks to its generative structure.
We demonstrated both theoretically and experimentally that the harmonium can extract multi-survival variate patterns --- such as three-way correlations --- that are impossible to discover with only one time-to-event variable. Furthermore, analysis of real-world data revealed that the harmonium captures information embodied in
complementary survival endpoints.
We have taken a first step in eliminating the
need for selecting a single endpoint and pave the way towards a unified timeline
view of the data.

\section*{Supporting information}

\paragraph*{S1 Appendix}
\label{S1_Appendix}
{\bf Appendix with supporting information.} Details of energy function, derivations of equations, and experimental aspects of datasets, model hyperparameters, and training.

\section*{Availability of data and materials}
Code, data, and examples are publicly available under the open source
\textit{Apache License 2.0} at
\url{https://gitlab.com/hylkedonker/harmonium-models}.

\section*{Acknowledgments}
We thank Rik Huijzer for proofreading the manuscript. We would like to thank the
Center for Information Technology of the University of Groningen for providing
access to the Peregrine high performance computing cluster.


\appendix
\section*{Appendix}
\section{Energy function survival variables}\label{sec:app_E_gamma}
All survival variables are assumed to be offset against a fixed landmark (e.g., the date when a participant entered the study) so that all values are $>0$ and we focus on right censored events.
Our goal is to construct an energy function for the survival variables  where the building blocks are composed of gamma distributions.
To this end, we make the following \textit{ansatz} for the energy function

\begin{equation}\label{eq:E_gamma}
    E_{B}(\bm{x}_B, \bm{h}) = \sum_{i \in B}\sum_{j \in H} x_i (W_B)_{ij} h_j  - \ln (x_i) |V_{ij}| h_j + \sum_{i \in B} x_i (a_B)_i - \ln (x_i) |c_i|.
\end{equation}

Henceforth $x_i$ are assumed to be scaled to the unit interval $x_i \in (0,1]$ for all $i\in B$ by normalising $x_i \rightarrow x_i/\tau_i$ with a suitably chosen time horizon $\tau_i$. Our motivation is two fold. The first is technical: $x_i \in (0,1]$ ensures that the probability density functions [Eq.~(\ref{eq:right_truncated_gamma}), below] can be normalised without imposing the constraint $\beta_i > 0$ for $\beta_i = \sum_{j \in H} (W_B)_{ij} h_j + (a_B)_i$ (to prevent diverging integrals near infinity). The second is physical: normalising the data reflects one's believe about the possible values the data can take since censoring at $\xi_i$ means: the actual event is supposed to occur somewhere in the interval $[\xi_i, 1]$.
For example, it may be unrealistic to assume that a person becomes over 120 years old, and the time horizon $\tau_i$ is a way to factor in these physical constraints.

Invoking the definition of the model $p(\bm{x}_B,\bm{h}) \propto \exp[-E_B(\bm{x}_B,\bm{h})]$ and normalising w.r.t. $\bm{x}_B$ yields the right truncated gamma distribution
\begin{equation}\label{eq:right_truncated_gamma}
    p_{\Gamma}\left(x_{i}|\bm{h}\right) = \frac{x_i^{\alpha_i - 1} e^{-\beta_i x_i}}{\Gamma(\alpha_i) \gamma^*(\alpha_i, \beta_i)} ,
\end{equation}
as the conditional probability distribution, where $\alpha_i = \sum_{j \in H} |V_{ij}|h_j + |c_i| + 1$, $\Gamma(x)$ the Gamma  function, and
\begin{equation}
    \gamma^*(a, z) = \frac{1}{\Gamma(a)} \int_0^1 t^{a-1} e^{-zt} \mathrm{d}t,
\end{equation}
the incomplete gamma function~\cite{PARI10}. Note that we haven't exhausted the entire parameter space by choosing the coupling strength $|V_{ij}|$ and bias $|c_i|$ to be positive, whence $\alpha_i \geq 1$. Technically, $\alpha_i$ must be larger than 0 to prevent poles from emerging.
By introducing a term $|c_i| \rightarrow |c_{i}| + d_i$ with $d_i > -1$ (e.g., $d_i=\lim_{A \uparrow 1} A \cos \varphi_i$) we can cover the entire domain of $\alpha_i$. But to simplify the generation of samples from Eq.~(\ref{eq:right_truncated_gamma}) we focus on the form laid out in Eq.~(\ref{eq:E_gamma}).
Observe that the bias $\bm{c}$ can in principle be captured by $\bm{V}$ at the expense of introducing additional latent states that are always turned on $h_i=1$. To reduce the amount of parameters as much as possible, we choose instead to model the bias $\bm{c}$ separate from $\bm{V}$.

\section{Derivation log-likelihood gradient}\label{app:ll_gradient}
The derivation presented here parallels Ref.~\cite{TRAN13} with the appropriate changes to the notation, and is provided here for completeness.
Our goal is to calculate the gradient of the log likelihood
\begin{equation}
    \mathcal{L}(\{\bm{o}^{(i)}\}_{i=1}^m) = \frac{1}{m}\sum_{i=1}^m \ln \int \mathrm{d}\bm{x} p(\bm{x}) \chi(\bm{x}, \bm{o}^{(i)}),
\end{equation}
(abbreviated using $\int \mathrm{d}\bm{x} \equiv \int_{-\infty}^{\infty} \mathrm{d}\bm{x}_C \int_0^1 \mathrm{d}\bm{x}_B \sum_{\bm{x}_A \in  \{0, 1\}\otimes^{|A|}}$) w.r.t. the expanded model
\begin{equation}\label{eq:tran_full}
    p(\bm{o}, \bm{x}, \bm{h}) = \frac{\exp[-E(\bm{x}, \bm{h})]\chi(\bm{x}, \bm{o})}{\Xi},
\end{equation}
where $\Xi$ is an unimportant normalisation constant which differs from the partition function $Z= \sum_{\bm{h}} \int  \mathrm{d}\bm{x} \exp[-E(\bm{x}, \bm{h})]$. To this end, write $Z(\bm{o}) \equiv \sum_{\bm{h}} \int \mathrm{d}\bm{x} \exp[-E(\bm{x}, \bm{h})] \chi(\bm{x}, \bm{o})$ to further simplify the log likelihood to
\begin{equation}
    \mathcal{L}(\{\bm{o}^{(i)}\}_{i=1}^m) = \frac{1}{m} \sum_{i=1}^m \ln \frac{Z\left(\bm{o}^{(i)}\right)}{Z}.
\end{equation}

To calculate $\nabla_{\bm{\Theta}} \mathcal{L}$ let us first compute $\nabla_{\bm{\Theta}} \ln Z(\bm{o})$. Working out the derivative of the first term
\begin{equation}\label{eq:nabla_Z_o}
    \nabla_{\bm{\Theta}} \ln Z(\bm{o}) = \frac{1}{Z(\bm{o})} \sum_{\bm{h}} \int \mathrm{d}\bm{x} \left[-\nabla_{\bm{\Theta}}E(\bm{x}, \bm{h}) \right] e^{-E(\bm{x}, \bm{h})} \chi(\bm{x}, \bm{o}).
\end{equation}
Substituting Eq.~(\ref{eq:tran_full}) with $p(\bm{o}) \equiv \sum_{\bm{h}} \int  \mathrm{d}\bm{x}p(\bm{o}, \bm{x}, \bm{h})$ in Eq.~(\ref{eq:nabla_Z_o})
\begin{equation}
    \frac{e^{-E(\bm{x}, \bm{h})} \chi(\bm{x}, \bm{o})}{Z(\bm{o})} = \frac{p(\bm{o}, \bm{x}, \bm{h})}{p(\bm{o})} = p(\bm{x}, \bm{h}|\bm{o}),
\end{equation}
shows that the normalisation constant $\Xi$ of Eq.~(\ref{eq:tran_full}) cancels out exactly, and therefore
\begin{equation}
    \nabla_{\bm{\Theta}} \ln Z(\bm{o}) = -\langle \nabla_{\bm{\Theta}}E(\bm{x}, \bm{h}) \rangle_{p(\bm{x}, \bm{h}|\bm{o})} \, .
\end{equation}
For the data-independent term $Z= \sum_{\bm{h}} \int  \mathrm{d}\bm{x} \exp[-E(\bm{x}, \bm{h})]$, we obtain the standard result~\cite{GOOD16}
\begin{equation}
    \nabla_{\bm{\Theta}} \ln Z = -\langle \nabla_{\bm{\Theta}}E(\bm{x}, \bm{h}) \rangle_{p(\bm{x}, \bm{h})} .
\end{equation}
Putting the two terms together, we arrive at the desired result
\begin{equation}
    \nabla_{\bm{\Theta}} \mathcal{L} = -\left( \frac{1}{m} \sum_{i=1}^m \left \langle \nabla_{\bm{\Theta}}E(\bm{x}, \bm{h}) \right\rangle_{p(\bm{x}, \bm{h}|\bm{o}^{(i)})} - \langle \nabla_{\bm{\Theta}}E(\bm{x}, \bm{h}) \rangle_{p(\bm{x}, \bm{h})} \right).
\end{equation}

\section{Synthetic two-dimensional survival distribution}\label{app:xor_dataset}
Blobs are two-dimensional independent (i.e., $t_1 \perp t_2$),
unit-interval truncated Gamma distributions
    [Eq.~(\ref{eq:p_interv_trunc_gamma})] with modes placed at
$\bm{v}^{(1)}=\left(\frac{1}{4},\frac{1}{4}\right)$, $\bm{v}^{(2)} =
    \left(\frac{3}{4},\frac{3}{4}\right)$ corresponding to red ($x_A=0$) and
$\bm{v}^{(3)}=\left(\frac{3}{4},\frac{1}{4}\right)$, $\bm{v}^{(4)} =
    \left(\frac{1}{4},\frac{3}{4}\right)$ for blue ($x_A=1$). Modes are sufficiently
squeezed (shape and rate $\alpha=8.1$, $\beta=58$ or $\alpha=29$, $\beta=76$) so
as to form a Gaussian-like shape and sampled with equal probability.

\section{Derivation harmonium as a mixture of Gaussians}\label{app:xor}
On a
high level, the event time density has four temporal modes located at
$\bm{v}^{(1)},\dots,\bm{v}^{(4)}$: two corresponding to binary colour $(x_A)_1=0$ ($\bm{v}^{(1)}$
and $\bm{v}^{(2)}$) and two for colour $(x_A)_1=1$ ($\bm{v}^{(3)}$ and $\bm{v}^{(4)}$).
Since there are no continuous variables $\bm{x}_C$ we
disregard corresponding terms in $E$, so that our goal will
be to compute $p(\bm{x}_A, \bm{x}_B) = \sum_{\bm{h}} \exp[-E(\bm{x}_A, \bm{x}_B, \bm{h})]/Z$ with
weights that fit the distribution.
We therefore allocate one hidden unit $h_i$ for each mode $\bm{v}^{(i)}$.
For convenience, write $x_A \equiv (x_A)_1$ since there is only one binary variable (colour).
Simplifying, by setting
the visible biases to zero ($\bm{c}=\bm{a}_A = \bm{a}_B=0$), we can evaluate
$p(x_A, \bm{x}_B$) up to a normalisation constant $Z$ by marginalising out $\bm{h}$:
\begin{equation}
    p(x_A, \bm{x}_B) = \sum_{\bm{h}} p(x_A, \bm{x}_B, \bm{h}) \propto \sum_{\bm{h}} \exp[-E(x_A, \bm{x}_B, \bm{h})] = \prod_{j=1}^4 1+e^{-\phi_j(x_A, \bm{x}_B)},
\end{equation}
where $\phi_j(x_A, \bm{x}_B)$ [Eq.~(12), Main Text] groups energy terms proportional
to latent state $h_j$.  Simplifying further, we substitute receptive fields $\bm{V}$ and
$\bm{W}_B$ in terms of its shape $\bm{\alpha} = |\bm{V}| + 1$ and rate
$\bm{\beta}=\bm{W}_B$, and replace $({W}_A)_{1j}$ by $(w_A)_j$ for
notational convenience. In this notation, we have
\begin{equation}\label{eq:app_phi}
    \phi_j = -\ln (\bm{x}_B)^T (\bm{\alpha}_j-1) + \bm{x}_B^T \bm{\beta}_j + (w_A)_j x_A + b_j,
\end{equation}
where we used $\bm{\alpha}_j$ to denote column $j$ of matrix $\bm{\alpha}$, and
similarly for $\bm{\beta}$. To pin $x_A$ to its corresponding  value $(\tilde{x}_A)_j$
of mode $j$, let $(w_A)_j=-q[(\tilde{x}_A)_j-\frac{1}{2}]$ and recall the Le Roux-Bengio
Kronecker delta identity~\cite{ROUX08}:
\begin{equation}\label{eq:app_kronecker}
    \lim_{q \rightarrow \infty} \exp\left\{-[(w_A)_j x_A - (w_A)_j (\tilde{x}_A)_j]\right\} = \delta_{x_A, (\tilde{x}_A)_j},
\end{equation}
when both $x_A$ and $(\tilde{x}_A)_j$ are binary valued.	Next, observe that most of the temporal mode's weight are concentrated around its maximum $v^{(j)}_i = (\alpha_{ij}-1)/\beta_{ij}$, justifying a Taylor expansion around it:
\begin{equation}\label{eq:app_taylor}
    (\alpha_{ij} - 1) \ln (x_B)_i  - \beta_{ij} (x_B)_i \approx (\alpha_{ij} - 1)\left[\ln v_i^{(j)} - 1 \right] - \frac{\beta_{ij}}{v^{(j)}_i} \frac{((x_B)_i - v_i^{(j)})^2}{2} + \mathcal{O}[((x_B)_i - v_i^{(j)})^3].
\end{equation}
With both identities [Eqs.~(\ref{eq:app_kronecker}) and (\ref{eq:app_taylor})] in hand, sweep the constants in the bias term:
\begin{equation}\label{eq:app_index}
    b_j = -(w_A)_j (\tilde{x}_A)_j + \sum_{i=1}^2 (\alpha_{ij} - 1)\left[\ln v_i^{(j)} - 1 \right] + \Lambda,
\end{equation}
together with a convergence factor $\Lambda$. Substituting Eqs.~(\ref{eq:app_taylor}) and (\ref{eq:app_index}) in (\ref{eq:app_phi}), and using identity (\ref{eq:app_kronecker}) we have:
\begin{equation}
    e^{-\phi_j} \approx \delta_{x_A, (\tilde{x}_A)_j} e^{-\Lambda} \exp \left[-\frac{1}{2} (\bm{x}_B-\bm{v}^{(j)})^T \bm{\Sigma}_j^{-1} (\bm{x}_B-\bm{v}^{(j)}) \right],
\end{equation}
with $\Sigma_j = \mathrm{diag}\left[\frac{\alpha_{1j}-1}{\beta_{1j}^2},
        \frac{\alpha_{2j}-1}{\beta_{2j}^2}\right]$ a diagonal covariance matrix.
Assuming that there is little overlap between the modes, i.e., $e^{-\phi_j}
    e^{-\phi_k} \approx 0$ for $j \neq k$, we have
\begin{equation}
    \tilde{p}(x_A, \bm{x}_B) = \prod_{j=1}^4 1+e^{-\phi_j} \approx 1 + \sum_{j=1}^4 e^{-\phi_j} = 1 + 2\pi e^{-\Lambda} \sum_{j=1}^4 \delta_{x_A,(\tilde{x}_A)_j}  \sqrt{|\bm{\Sigma}_j|} \mathcal{N}\left(\bm{v}^{(j)}, \bm{\Sigma}_j \right),
\end{equation}
where $|\bm{\Sigma}_j|$ is used to denote the determinant of covariance matrix $\bm{\Sigma}_j$.
Finally, assume that each Gaussian $\mathcal{N}\left(\bm{v}^{(j)}, \bm{\Sigma}_j \right)$ is sufficiently localised on the $[0,1]\times [0,1]$ unit square so that
\begin{equation}
    \int_0^1 \mathrm{d}(x_B)_1 \int_0^1 \mathrm{d}(x_B)_2 \, \mathcal{N}\left(\bm{v}^{(j)}, \bm{\Sigma}_j \right) \approx \int_{-\infty}^{\infty} \mathrm{d}(x_B)_1 \int_{-\infty}^{\infty} \mathrm{d}(x_B)_2 \, \mathcal{N}\left(\bm{v}^{(j)}, \bm{\Sigma}_j \right) = 1.
\end{equation}
This integral identity allows us to normalise $\tilde{p}(x_A, \bm{x}_B)$
\begin{equation}
    \sum_{x_A \in \{0,1\}} \int_0^1 \mathrm{d}(x_B)_1 \int_0^1 \mathrm{d}(x_B)_2 \, \tilde{p}(x_A, \bm{x}_B) \approx 2 + 2\pi e^{-\Lambda}\sum_{j=1}^4  \sqrt{|\bm{\Sigma}_j|},
\end{equation}
giving rise to the overall solution as a mixture of Gaussians:
\begin{equation}
    p(x_A, \bm{x}_B) \approx \sum_{j=1}^4 \pi_j \delta_{x_A,(\tilde{x}_A)_j} \mathcal{N}\left(\bm{v}^{(j)}, \bm{\Sigma}_j\right),
\end{equation}
with weights $\pi_j = \frac{\sqrt{|\bm{\Sigma}_j|}}{\sum_{k=1}^4 \sqrt{|\bm{\Sigma}_k|}}$ after choosing a sufficiently large negative convergence factor $\Lambda$. Finally, substituting $\bm{W}_B$ and $\bm{V}$ back into $\bm{v}^{(j)}$ and $\bm{\Sigma}_j$ we arrive at the mean and the (diagonal) covariance matrix in terms of the receptive fields:
\begin{equation}
    v_i^{(j)} = \frac{|V_{ij}|}{(W_B)_{ij}}, \quad (\Sigma_{ii})_j=\frac{|V_{ij}|}{(W_B)_{ij}^2}.
\end{equation}

\section{Experimental aspects}\label{app:experimental}
\subsection{Metrics}\label{sec:app_metrics} The concordance
index~\cite{HARR96} orders the data according to the event time, and measures
the amount of data pairs in which the model's risk prediction is ordered
concordantly. The concordance index is thus independent of the exact risk scores
but only measures their relative ranking. We therefore chose a fixed time point
$t$ at half the time horizon $t=\tau/2$, and defined the risk score as the predicted survival at that time point
i.e.,
\begin{equation}\label{eq:risk}
    r_i=S\left(x_i=t|\bm{o}_{-i}\right) = \frac{p\left(x_i > t,\bm{o}_{-i}\right)}{p(\bm{o}_{-i})},
\end{equation}
where $\bm{o}_{-i}$ denotes observation $\bm{o}$ with element $i$ removed. In
addition, we used $r_i$ to compute the Brier loss~\cite{GRAF99} to measure the
calibration at time point $t$. Notice that Eq.~(\ref{eq:risk}) factors in the
survival information from all other survival variables, when there is more than
one time-to-event variable. The risk score $r_i$ when marginalised over survival
variable $j$ is obtained by censoring at time zero [i.e., $o_j = (\xi_j=0, e_j=0)$] so that
$r_i=S(x_i=t|\bm{o}_{- \{i,j\}})$. Observe, moreover, that the right hand side of
Eq.~(\ref{eq:risk}) can be evaluated in terms of its unnormalised probabilities since
the partition function cancels out.

There are atleast two ways to compute $r_i$ via the unnormalised probability
density $\tilde{p}(\bm{o}, \bm{x}, \bm{h}) = e^{-E(\bm{x}, \bm{h})}\chi(\bm{x},
    \bm{o})$: (i) integrate out $\bm{x}$ analytically and then sum over $\bm{h}$
numerically or (ii) carry out the $\bm{h}$ sum analytically and numerically
marginalise over $\bm{x}$. While the computational complexity of the former
method is linear in the number of visible units $n_v$ and exponential in the
number of latent states $n_h$, the latter scales linearly in $n_h$ and roughly
exponentially in the number variables with censored/missing values. We therefore
used, for the datasets presented here, method (i) when $n_h < 10$ and method
(ii) otherwise.

\subsection{Generation of samples}
To sample from Eqs.~(10,11,17) Main Text, requires samples from the sigmoid function, Gaussian distribution, right truncated Gamma distribution and the interval truncated Gamma distribution. Gaussian samples can be generated using the SciPy routine and binary states can be sampled by picking 1 when the sigmoid activation function exceeds a $[0,1]$ uniformly sampled threshold, and 0 otherwise. To sample from the $[t_<,1]$ interval truncated Gamma distribution
\begin{equation}\label{eq:p_interv_trunc_gamma}
    p_{[t_<,1]}^\Gamma(x|\alpha, \beta) =
    \frac{x^{\alpha-1}e^{-\beta x}}{\int ^1_{t_<}\mathrm{d}t \, t^{\alpha-1}e^{-\beta t}}
    = \frac{\theta(x - t_<)}{1 - t_<^{\alpha} \frac{\gamma^*(\alpha, t_< \beta)}{\gamma^*(\alpha, \beta)}} p_{\Gamma}\left(x|\alpha, \beta\right),
\end{equation}
and the right truncated Gamma distributions $p_{\Gamma}\left(x|\alpha, \beta \right)$ [Eq.~(\ref{eq:right_truncated_gamma})], observe that the samples from the latter can be obtained from the former with $t_<=0$. Unfortunately, we are not aware of any existing algorithms that can generate samples from intervals for $\alpha >0$ and both $\beta \geq 0$ and $\beta < 0$.
By noticing that
\begin{equation}
    x^{\alpha-1}e^{-\beta x} \leq e^{(\alpha-1-\beta) x} \frac{1}{\exp(\alpha-1)} ,
\end{equation}
for $\alpha > 1$, we propose the following rejection algorithm $\forall \beta$ and $\alpha > 1$:

\begin{algorithm}[H]
    \label{alg:truncated_gamma}
    \begin{algorithmic}[1]
        \WHILE{True}
        \STATE sample $u \sim U_{[0,1]}$ and $y \sim c_{[0, 1-t_<]}(y|\alpha - 1 - \beta)$
        \STATE compute $p_\mathrm{accept}(x)$ with $x = 1 - y$.
        \IF{$u \leq p_\mathrm{accept}$}
        	\RETURN $x$\;
        \ENDIF
        \ENDWHILE
    \end{algorithmic}
    \caption{Sampling method for the $[t_<, 1]$ interval truncated gamma distribution for $\alpha>1$.}
\end{algorithm}
with
\begin{equation}\label{eq:p_accept_algo1}
    p_\mathrm{accept}(x) = \left(\frac{x}{\exp(x-1)}\right)^{\alpha -1},
\end{equation}
$U_{[0,1]}$ the uniform distribution on the unit interval, and $c_{[0,t]}(x|\lambda)$ the exponential decaying distribution
\begin{equation}\label{eq:exp_truncated}
    c_{[0,t]}(x|\lambda) = \frac{\lambda e^{-\lambda x}}{1 - e^{-t\lambda}},
\end{equation}
normalised on the interval $[0,t]$. Samples from Eq.~(\ref{eq:exp_truncated}) can be generated by inverting its cumulative distribution $\mathcal{C}(x|\lambda,t) = (1-\exp[-\lambda x])/(1-\exp[-t\lambda])$.

\subsection{Initialisation of parameters}
For the categorical weights (in $E_A$), $\bm{W}_A$ was drawn from the Gaussian $\mathcal{N}(0, 0.01)$ and
the bias $\bm{a}_A$ was initialised to $\ln [(1-\bm{p}) \oslash \bm{p}]$ where $\bm{p}$ is the corresponding average value of the categorical variable in the training set (ignoring any missing values),
as suggested in Ref.~\cite{HINT12}.
Initialisation of the parameters in $E_C$ followed Ref.~\cite{MELC17} by using Glorot-Bengio samples~\cite{GLOR10} $\bm{W}_C \sim \left[-\sqrt{\frac{6}{|H| + |C|}}, \sqrt{\frac{6}{|H| + |C|}}\right]$ and $\bm{\sigma}$ was treated as an adjustable parameter with initial value 1.
The bias $\bm{a}_C$ was set to zero since the input features can be standardised prior to training.
Similarly, for the time-to-event parameters (in $E_B$) we used $\bm{W}_B \sim \left[-\sqrt{\frac{6}{|H| + |B|}}, \sqrt{\frac{6}{|H| + |B|}}\right]$ and sampled both $\bm{V}$ and $\bm{c}$ uniformly from $\left[0, 2\sqrt{\frac{6}{|H| + |B|}}\right]$ to ensure unit variance~\cite{GLOR10}, and picked $\bm{a}_B=0$.
Finally, hidden biases were set to $\bm{b}=0$ as recommended in~\cite{HINT12}.

\subsection{Encoding experimental datasets}
In this section, we indicate what variables were used from the datasets and how they were transformed.

\begin{itemize}
    \item \textit{arrest}: All features were used for training, but we grouped
          the number of prior convictions $>$ 5 before one-hot encoding categories as dummies.

    \item \textit{democracy}: We only considered the continent
          name and type of regimes as features.
    \item \textit{gbsg2}: All features were used.

    \item \textit{ncctg}: All features except
          for the institute code and the columns weight loss and meal calory intake were used.
          After one-hot encoding caterories as dummies, the low variance features that were on/off in more
          than 95 \% of the samples were dropped to prevent collinearity.
    \item \textit{nvalt11}: We modelled the categoric variables gender,
          control arm, performance status, and smoking status plus the numeric feature
          age. The categorical features histology, prior medical conditions, prior
          malignancies, and stage and numeric variable BMI contained missing values, and
          were therefore dropped in all models except the harmonium.

    \item \textit{nvalt8}: All models used the numeric feature age and the categories: performance
          status, histology, smoking status, stage of the disease, control arm and the
          T, N, and M tumour classification categories. After one-hot-encoding, the low
          variance features that were on/off in more than 95 \% of the samples were
          dropped. The harmonium also incorporated
          the metabolic activity measured as FDG-PET SUV$_{\mathrm{max}}$ $\geq$10 and
          the numeric variable BMI that both contained missing values.
\end{itemize}

Apart from the time-to-event variables, all numeric features were standardised and categorical variables were dummy encoded prior to training.

\subsection{Settings of benchmark real world datasets}
The concordance index~\cite{HARR96} and the Brier loss~\cite{GRAF99} (which was
not available for the SVM) were measured using 5x5 nested
cross-validation~\cite{STON74} where the inner loop was used to hyperparameter
tune the model with the random search algorithm~\cite{BERG12} from
Scikit-learn~\cite{PEDR11} using 50 samples. Since both \textit{nvalt} datasets
consists of two time-to-event variables while the Cox model, SVM, and random
forest can only consider a single time-to-event variable, we chose to train and
evaluate these models on the OS while the harmonium was trained on both survival
variables and  was evaluated on the OS. The Brier loss was computed at
$\frac{1}{2}\tau_{\mathrm{OS}}$ with $\bm{\tau}$ the time horizon, which for each survival variable was set to the largest time recording in the dataset. For the harmonium, the same time point $\tau_{\mathrm{OS}}/2$ were used for
computing risk scores (see~\ref{sec:app_metrics}). Hyperparameters were tuned to optimise the
concordance index, where the harmonium factored in the RFS and SBMFS variable to
predict OS in the \textit{nvalt8} and \textit{nvalt11} dataset, respectively.

For the Cox model from lifelines~\cite{DAVI20}, the regularisation term $R(\bm{\beta})$ is parametrised as
\begin{equation}
    R = \frac{\lambda_C}{2}\left[ (1-\ell_1) \|\beta\|^2_2 + \ell_1 \|\beta\|_1\right],
\end{equation}
where $\bm{\beta}$ are the coefficients of the model. Parameters $\lambda_C$ and
$\ell_1$ were sampled log-uniformly from the intervals $[10^{-5}, 10^3]$ and
$[10^{-5}, 1]$, respectively.

For the survival SVM, the hyperparameter of the squared Hinge loss $\alpha$ was
sampled log uniformly from $[2^{-12}, 2^{12}]$ while the ranking ratio $r$ was
uniformly sampled from $[0,1]$ in steps of 0.05, as suggested in
Ref.~\cite{POLS15}.

For the random survival forest, we selected a maximum tree depth of 7 (instead
of unbounded) to reduce the memory footprint, and varied (i) the number of
estimators as $2^j$ uniformly from $j=0,\dots,10$, (ii) the minimum of samples
required for a split as $2^k$ uniformly from $k=1,\dots,5$, (iii) the minimum
number of samples per leaf as $2^l$ uniformly from $l=0,\dots,5$ and, (iv)
maximum number of features to consider per split by randomly selecting any of
$\sqrt{n}$, $\log_2 n$, or $n$ with equal probability, with $n$ the number of
features.

Finally, for the harmonium (i) the number of hidden units, (ii) learning rate,
(iii) number of epochs to train, (iv) mini batch size, and (v) $L_2$ penalty
$R(\bm{\Theta})=\lambda_H/2 \Theta^2$ were all sampled log uniformly from
$[1,128]$, $[10^{-5}, 5\cdot 10^{-2}]$, $[500, 10^5]$, $[25, 10^3]$, and
$[10^{-5}, 10^{-1}]$, respectively. In each gradient step, a part of the
previous update was retained using a momentum fraction $1-f$, where $f$ was
chosen uniformly from $[0, 0.9]$. And lastly, we allowed the Gibbs chain of the
negative phase to persist~\cite{TIEL08} instead of re-initialising it each step
as in Algorithm 1, Main Text. We considered this as an hyperparameter as
well, and chose either options with 50 \% chance, and fixed the number of
contrastive divergence steps to 1. The following exceptions were made to these settings:
(i) for the \textit{democracy} and \textit{ncctg} dataset the number of epochs was capped to
$5\cdot10^4$ to reduce the computation time, (ii) and we lowered the maximum
learning rate for the \textit{nvalt8} dataset to 0.0125 to prevent numerical
instability.

\bibliographystyle{unsrt}
\bibliography{donker_arxiv}

\end{document}